\documentclass{article}

\usepackage{microtype}
\usepackage{graphicx}
\usepackage{subcaption}
\usepackage{booktabs}
\usepackage{hyperref}

\usepackage[accepted]{icml2026}

\usepackage{amsmath}
\usepackage{amssymb}
\usepackage{mathtools}
\usepackage{amsthm}

\usepackage[capitalize,noabbrev]{cleveref}

\theoremstyle{plain}

\theoremstyle{definition}

\theoremstyle{remark}

\newcommand{\methodName}{OGD}

\usepackage{enumitem}

\icmltitlerunning{Ontology-Guided Diffusion for Zero-Shot Visual Sim2Real Transfer
}

\begin{document}

\twocolumn[
  \icmltitle{Ontology-Guided Diffusion for Zero-Shot Visual Sim2Real Transfer}

  \begin{icmlauthorlist}
    \icmlauthor{Mohamed Youssef}{vis,ipa}
\icmlauthor{Mayar Elfares}{vis}
\icmlauthor{Anna-Maria Meer}{ipa}
\icmlauthor{Matteo Bortoletto}{vis}
\icmlauthor{Andreas Bulling}{vis}
  \end{icmlauthorlist}

  \icmlaffiliation{vis}{University of Stuttgart, Germany}
  \icmlaffiliation{ipa}{Fraunhofer IPA, Germany}

  \icmlcorrespondingauthor{Mayar Elfares}{mayar.elfares@vis.uni-stuttgart.de}

  \icmlkeywords{Machine Learning, Diffusion Models, Knowledge Graphs, Sim2Real}

  \vskip 0.3in
]

\printAffiliationsAndNotice{}

\begin{abstract}
Bridging the simulation-to-reality (sim2real) gap remains challenging as labelled real-world data is scarce. Existing diffusion-based approaches rely on unstructured prompts or statistical alignment, which do not capture the structured factors that make images look real. We introduce \textit{Ontology-Guided Diffusion (\methodName{})}, a neuro-symbolic zero-shot sim2real image translation framework that represents realism as structured knowledge. \methodName{} decomposes realism into an ontology of interpretable traits – such as lighting and material properties – and encodes their relationships in a knowledge graph. From a synthetic image, OGD infers trait activations and uses a graph neural network to produce a global embedding. In parallel, a symbolic planner uses the ontology traits to compute a consistent sequence of visual edits needed to narrow the realism gap. The graph embedding conditions a pretrained instruction-guided diffusion model via cross-attention, while the planned edits are converted into a structured instruction prompt. Across benchmarks, our graph-based embeddings better distinguish real from synthetic imagery than baselines, and OGD outperforms state-of-the-art diffusion methods in sim2real image translations. Overall, OGD shows that explicitly encoding realism structure enables interpretable, data-efficient, and generalisable zero-shot sim2real transfer.
\end{abstract}


\section{Introduction}


\begin{figure}[t]
    \centering
    \includegraphics[width=\columnwidth]{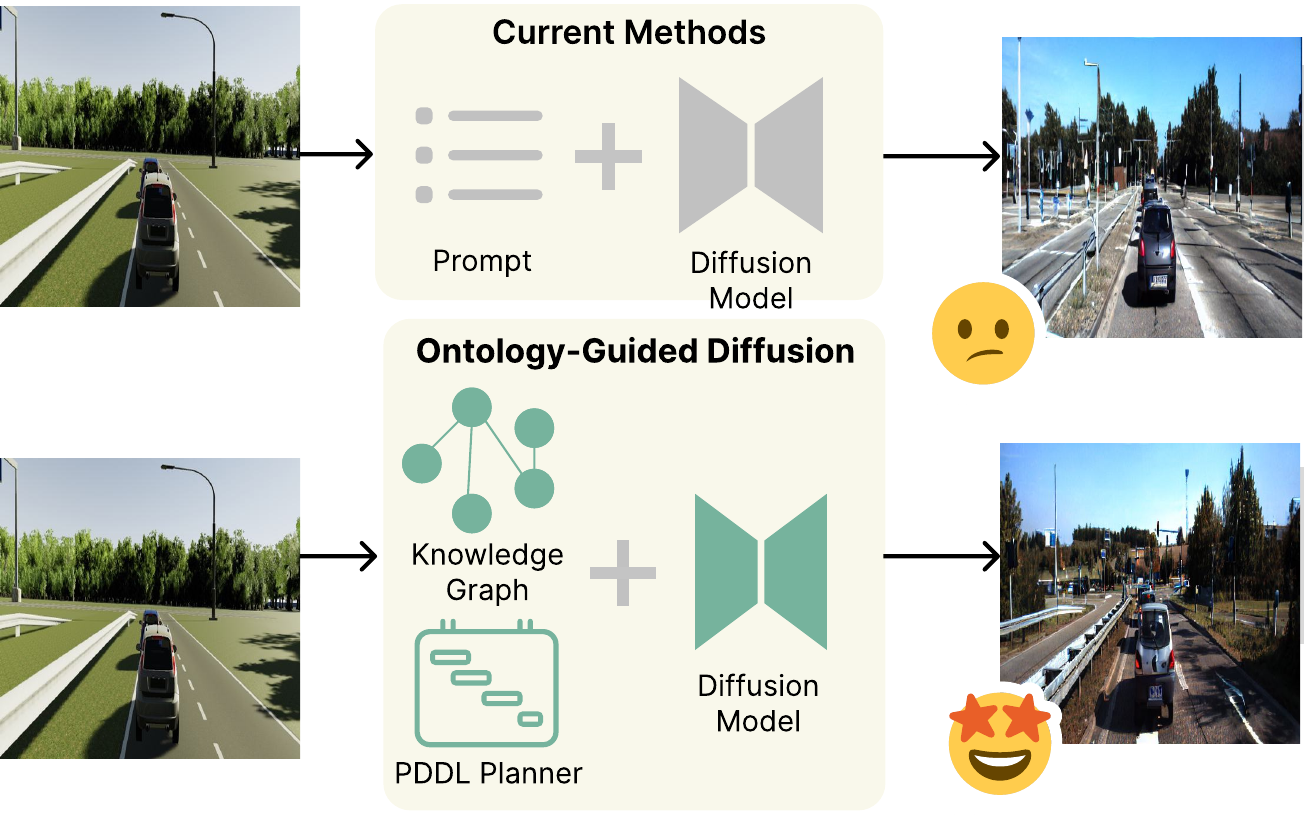}
    \caption{
    \textbf{Ontology-Guided Diffusion (\methodName{}) explicitly models visual realism to bridge the sim2real gap.} Unlike current instruction-guided diffusion that relies on unstructured prompts, \methodName{} decomposes realism into structured traits organised in a knowledge graph that models causal relationships, and uses a symbolic PDDL planner to generate coherent editing actions. Conditioning the diffusion model on this structured guidance produces more realistic and consistent translations from synthetic to real images, enabling interpretable and data-efficient sim2real transfer.
  }
  \label{fig:teaser}
\end{figure}

\begin{figure*}[t]
  \centering
  \includegraphics[width=2\columnwidth]{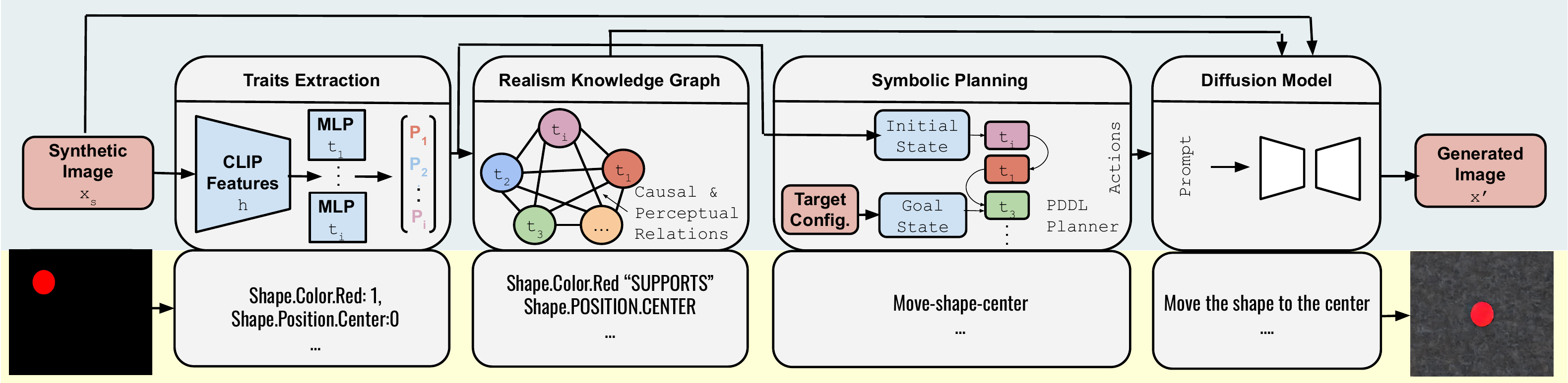}
  \caption{
    Overview of the proposed ontology-guided diffusion framework.
    A synthetic image is first mapped to realism trait probabilities using supervised MLP heads trained on frozen CLIP features.
    Traits are propagated through a static realism knowledge graph using a GNN to obtain node-level realism embeddings.
    Differences between synthetic and target realism states are converted into symbolic transformation plans via PDDL.
    The symbolic plan and graph embeddings jointly condition a diffusion-based image editing model.}
  \label{fig:pipeline}
\end{figure*}

Learning visual models in simulation and deploying them in the real world has long been hindered by the simulation-to-reality (sim2real) gap \cite{tobin2017domain, pitkevich2024survey}. 
While modern simulators can generate large volumes of labelled data at low cost, synthetic images often fail to capture complex visual cues present in real-world imagery, including lighting irregularities \cite{bai2024close}, material imperfections \cite{jianu2022reducing}, sensor artefacts \cite{mahajan2024quantifying}, or subtle scene interactions \cite{tobin2017domain}. 
As a result, models trained purely on synthetic data frequently exhibit degraded performance when transferred to real environments, particularly in robotics and embodied perception settings \cite{sadeghi2017cad2rl}.

Recent progress in generative diffusion models has renewed interest in sim2real image translation as a means of narrowing this gap. Image-to-image diffusion frameworks \cite{Saharia2022Palette}, including instruction-guided editing models \cite{brooks2023instructpix2pix}, provide a flexible mechanism for transforming synthetic images to a more realistic appearance.

However, existing diffusion-based sim2real approaches typically rely on unstructured textual prompts or implicit statistical alignment between domains. 
In practice, such prompts can be ambiguous and hard to control, and they do not explicitly encode or reason about the causal structure underlying visual realism. 
As a consequence, these methods require extensive manual prompt engineering, large amounts of paired data, and domain-specific tuning.

A key limitation of current approaches is the absence of an explicit representation of \emph{what realism is}. 
Prior work in computer graphics and visual perception has established that visual realism is not a monolithic attribute, but arises from the interaction of multiple factors such as illumination gradients, shadow behaviour, material properties, optical distortions, and scene-level consistency~\cite{ferwerda2003three,carlson2018modeling}. 
Moreover, prior work has documented that these realism factors exhibit systematic dependencies imposed by physical and perceptual constraints: some cues reinforce one another, while others are mutually incompatible~\cite{ferwerda2003three,carlson2018modeling}. 
Despite this well-established structure, modern generative models largely ignore such knowledge and instead treat realism as an implicit, entangled latent variable.

In this work, we argue that sim2real transfer can be made more interpretable, controllable, and data-efficient by explicitly modelling key aspects of photorealism as a structured latent.
To demonstrate this, we introduce \emph{Ontology-Guided Diffusion (\methodName)}\footnote{We will make all the papers' artefacts public upon acceptance.}, a neuro-symbolic framework that integrates prior knowledge about visual realism directly into the conditioning of diffusion models, while leaving the base diffusion training procedure unchanged.
Rather than relying on unstructured textual prompts or 
entangled latent representations, our approach decomposes realism into a fixed ontology of visual traits grounded in established graphics and perception principles \cite{carlson2018modeling,ferwerda2003three}, encoding supportive and opposing relationships that capture causal and perceptual constraints between realism cues.

As shown in Figure \ref{fig:pipeline}, given an input image, we first infer trait probabilities using a CLIP-based multi-head classifier trained on frozen visual features. 
These activations are propagated through the knowledge graph using a graph neural network (GNN), producing a structured embedding that measures the global coherence of realism cues present in the image. 
To translate realism discrepancies into actionable guidance, we introduce a symbolic planning module based on the Planning Domain Definition Language (PDDL) \cite{mcdermott1998pddl}. 
The planner takes the predicted traits of a synthetic image and a target realism specification, and computes a minimal, causally consistent sequence of visual transformations required to bridge the gap. 
This plan is then converted into an instruction prompt that replaces unstructured natural language descriptions.
Both the realism graph embedding and the symbolic plan are integrated into an InstructPix2Pix diffusion model \cite{brooks2023instructpix2pix}. 
The graph embedding is injected as additional conditioning via cross-attention, providing the denoising network with access to information about the underlying structural relationships during generation.
The symbolic plan provides procedural guidance, specifying \emph{what} visual changes should occur and in \emph{which order}. 
During training, we further enforce realism through a latent alignment loss that encourages generated images to match target realism embeddings, rather than merely minimising pixel-level discrepancies.

We evaluate our approach along two complementary axes.
First, we assess the quality of the learned realism representations by benchmarking a meta-classifier trained on realism graph embeddings against baselines operating on raw visual features or unstructured trait vectors.
Second, we evaluate sim2real image translation performance, comparing \methodName{} against state-of-the-art diffusion baselines under zero-shot and low-data settings.
Our method consistently outperforms the state-of-the-art baselines (c.f. Section \ref{sec:experiments}).

Our results indicate that explicitly structured realism guidance not only improves both semantic realism alignment and perceptual image quality, but is also highly data-efficient: During training, it requires a small set of \textit{only} 140 \textit{unpaired} images for the meta-classifier by using structured semantic traits, as opposed to pixel-level classification models. At inference time, the approach operates in a zero-shot manner\footnote{Throughout this paper, we use the term zero-shot sim2real in the sense of zero-shot generalization at inference time, following the convention of large pretrained models such as SAM \cite{kirillov2023segment}.}, whereas competing methods typically require at least hundreds of domain-specific images for adaptation. This combination of minimal training data requirements and zero-shot inference makes our approach particularly well suited to robotics and simulation-heavy domains, where collecting paired synthetic–real data is costly or infeasible.

In summary, our work makes the following contributions:
\begin{itemize}[leftmargin = *]
    \item We introduce a novel ontology-based formulation of visual realism, modelling appearance realism as a structured set of interpretable traits with signed causal relationships grounded in graphics and perception literature.
    \item We propose a graph-based realism embedding that explicitly encodes global realism coherence and enables quantitative evaluation of realism structure beyond pixel-level similarity.
    \item We present \methodName{}, the first sim2real image translation framework that formulates realism transformation as a symbolic planning problem and integrates PDDL-generated action plans directly into diffusion-based image editing.
    \item We demonstrate that ontology-guided diffusion enables interpretable, data-efficient, and zero-shot sim2real transfer, consistently outperforming diffusion baselines that rely on unstructured prompt conditioning. 

\end{itemize}

\section{Related Work}
\paragraph{Sim-to-Real Transfer.}
The sim2real gap remains a key challenge in robotics and visual perception due to systematic differences between synthetic and real imagery, such as lighting, materials, and sensor artefacts \cite{carlson2018modeling, ferwerda2003three}. Early domain randomisation approaches improve robustness by introducing appearance variation \citep{tobin2017domain}, but often produce unrealistic images and lack an explicit notion of realism, limiting their effectiveness for perception tasks. Adversarial translation methods such as SimGAN \citep{shrivastava2017learning} and CycleGAN \citep{zhu2017unpaired} align synthetic and real distributions, but require large datasets and treat realism as an unstructured latent, reducing interpretability and generalisation. In contrast, our approach explicitly models realism as a structured, interpretable latent via an ontology of appearance-level traits, enabling semantically consistent sim2real translation with minimal data and robust zero-shot transfer at inference time.

\paragraph{Diffusion-Based Sim-to-Real Methods.}
Diffusion models have recently emerged as a robust alternative to GAN-based sim2real methods, offering more stable training and higher-fidelity image generation \citep{ho2020ddpm, rombach2022latent}. Image-to-image diffusion frameworks enable controlled visual transformations and have been applied to sim2real transfer via conditional generation, for example, through layout-conditioned models that enhance synthetic scenes using spatial constraints \citep{li2024aldm}. However, most existing diffusion-based approaches remain largely statistical, relying on paired or weakly paired data and unstructured conditioning signals, which leaves visual realism implicitly learned and limits interpretability, controllability, and data efficiency. In contrast, our approach explicitly integrates structured realism knowledge into the diffusion process by conditioning generation on symbolic transformation plans and graph-based realism embeddings, enabling semantically grounded, causally consistent sim2real transfer with reduced reliance on large paired datasets.

\paragraph{Structured Conditioning and Intermediate Representations.}
Prior work uses structured intermediate representations such as semantic layouts, depth maps, and surface normals to guide visual generation \citep{zhang2023controlnet}, improving controllability but relying on geometry- or simulator-specific annotations that limit their applicability to sim2real transfer. In contrast, our approach is appearance-level and simulator-agnostic: realism traits are defined in terms of visual phenomena rather than scene geometry or simulator internals, allowing reuse across different simulators and real-world datasets. While natural-language conditioning has also been explored as a domain-bridging signal \citep{yu2024language}, it is often too coarse to capture fine-grained appearance factors critical for perceptual realism. Our method instead introduces a structured, interpretable representation that explicitly encodes such appearance-level realism factors and their dependencies, complementing diffusion-based generation.

\paragraph{Knowledge Graphs and Symbolic Reasoning.}
Knowledge graphs and ontologies have been widely used to encode structured priors in vision and robotics, supporting explicit relational reasoning \cite{tenorth2013knowrob, battaglia2018relational}, and have been applied to generative modeling via scene graph–to–image synthesis and graph-conditioned diffusion models \citep{johnson2018image, menneer2025hig}. However, these approaches primarily model semantic scene structure, such as object identities and spatial relations, rather than appearance-level factors that govern visual realism. In contrast, our work introduces a realism-specific ontology grounded in graphics and perception, encoding causal relationships among lighting, materials, shadows, and sensor artefacts, and integrates symbolic planning via PDDL to produce interpretable, causally consistent transformation sequences for diffusion-based image editing.

\section{Method}
\label{sec:method}

We introduce Ontology-Guided Diffusion (\methodName), a neuro-symbolic framework for zero-shot sim2real image translation. The central idea is to explicitly model visual realism as a structured latent variable, rather than treating it as an implicit, entangled property learned end-to-end by a diffusion model.
To achieve this goal, our method decomposes sim2real translation into four stages, as shown in Figure \ref{fig:pipeline}.

We do not introduce new realism rules; instead, our ontology formalises existing constraints and appearance cues from vision and graphics literature \cite{ferwerda2003three, carlson2018modeling} into a signed knowledge graph, making their relationships explicit and machine-interpretable. Our contribution lies in converting these established rules (cf. a subset of the rules in Table~\ref{tab:trait_relations_vlm})
into a structured graph and integrating it into a diffusion-based sim2real pipeline. Experiments show that graph-based realism embeddings capture discriminative structure and separate real from synthetic imagery more effectively than unstructured traits or raw visual features.


Given a synthetic input image $x_s$, our objective is to generate an edited image $\hat{x}$ that preserves the semantic content of $x_s$ while matching the appearance-level realism of a target domain.
The target realism can be specified either implicitly via a real reference image $x_t$, or explicitly through a desired configuration of realism traits.


\subsection{Trait-Based Representation of Visual Realism}
\label{sec:traits}

We represent visual realism using a fixed set of $N$ appearance-level traits
$\mathcal{T} = \{t_1, \dots, t_N\}$  lighting, shadows, material properties,
geometric cues, optical and sensor artefacts, colour behaviour, and scene consistency.
The full list of traits, together with their definitions and supporting references
from graphics and perception literature, is provided in Appendix~\ref{app:ontology}.

\paragraph{Visual Feature Extraction.}
Given an image $x$, we extract a global visual representation using a frozen CLIP image encoder:
\begin{equation}
    \mathbf{h} = \mathrm{CLIP}(x) \in \mathbb{R}^d.
\end{equation}
The CLIP encoder is kept fixed throughout all experiments, providing stable, high-level
visual features learned from large-scale pretraining and avoiding task-specific overfitting.

\paragraph{Trait Prediction via Supervised Heads.}
Rather than relying on CLIP for zero-shot trait prediction, we train a set of lightweight,
supervised MLP classification heads $\{f_i\}_{i=1}^N$, one per realism trait.
Each head predicts the presence of a specific appearance-level property, given the shared
CLIP feature vector $\mathbf{h}$.
The predicted probability for the presence of the trait $t_i$ is given by:
\begin{equation}
p_i = \mathrm{softmax}(f_i(\mathbf{h}))_1,
\end{equation}
yielding a trait probability vector
$\mathbf{p} = (p_1, \dots, p_N) \in [0,1]^N$.
We employ a vision-language model (VLM), e.g., GPT-4o \cite{openai_gpt4o_2024}, as an automatic annotator to provide supervision for training realism trait predictors. The VLM analyzes each image and assigns labels indicating the presence or absence of predefined appearance-level traits, such as lighting effects, material properties, or sensor artifacts. These VLM-generated annotations serve as pseudo-labels for training the supervised MLP classification heads\footnote{Importantly, the VLM is not employed during inference and does not interact with the diffusion-based generation process; its role is strictly to provide scalable, label-efficient supervision in place of manual annotation. Any uncertain or missing trait predictions produced by the VLM are explicitly masked during training to prevent the introduction of noisy supervision.}.

Each realism trait is modeled as a binary classification problem with two output logits corresponding to absence and presence; we apply a softmax and take the probability of the positive (present) class as the trait activation.


\subsection{Realism Ontology and Knowledge Graph}
\label{sec:kg}\label{sec:gnn}

To encode prior knowledge about how realism traits interact, we define a static realism
knowledge graph $G = (V, E)$.
Each node $v_i \in V$ corresponds to a realism trait $t_i \in \mathcal{T}$ (c.f. Figure \ref{fig:pipeline}).
Edges encode causal or perceptual relationships between traits and are assigned signed
weights $w_{ij} \in [-1,1]$, where positive values indicate supportive relationships
(e.g., non-uniform lighting supports the presence of cast shadows), and negative values
indicate opposing relationships (e.g., perfectly sharp geometry conflicts with surface
imperfections).
The structure of the graph and the sign of each relationship are based on prior work that documents how visual realism arises from
interacting cues rather than independent factors, and how certain combinations of cues
are perceptually inconsistent
\cite{ferwerda2003three, carlson2018modeling}.
Table~\ref{tab:trait_relations_vlm} summarises the specific trait--trait relationships and
their supporting evidence from the literature.

The realism graph is fixed across all experiments: no edges, weights, or node definitions
are learned from data.
This design choice reflects our goal of injecting structured, domain-independent prior
knowledge into the model, rather than allowing the realism structure itself to overfit
to a specific dataset, enforcing our zero-shot generalisation.
We therefore treat the ontology as a domain-agnostic prior, while any domain-specific variation is captured only through the predicted trait activations.

\paragraph{Graph Realism Embedding}
Although the structure and edge weights of the realism knowledge graph are fixed a priori, a graph neural network (GNN) is trained on top of this static graph to produce embeddings that are consistent with the ontology. The GNN propagates information across the graph, allowing each node embedding to incorporate contextual knowledge from related traits while respecting the predefined relational structure. These resulting embeddings serve as a structured, semantically grounded representation of realism, which can then be injected as conditioning signals into the diffusion model. By integrating these ontology-consistent embeddings, the diffusion process is guided to produce images that adhere to the relational and perceptual constraints encoded in the knowledge graph, improving semantic fidelity, visual realism, and causal consistency in sim2real translations.

Technically, given an image-specific trait probability vector $\mathbf{p}$, we initialize node features as:
$\mathbf{x}_i = p_i$ and use a GraphSAGE-based GNN \cite{hamilton2017graphsage} to propagate information across the realism graph: $\mathbf{z}_i = \mathrm{GNN}(\mathbf{x}_i, G), \mathbf{z}_i \in \mathbb{R}^k$.
The loss used to train our GNN loss is:
\begin{equation}
\mathcal{L}_{\text{GNN}} =
\mathcal{L}_{\text{sim}} + \lambda_{\text{rep}} \mathcal{L}_{\text{rep}}.
\end{equation}
\begin{equation} \text{where, }
\mathcal{L}_{\text{sim}} =
\frac{1}{|E|}
\sum_{(i,j)\in E}
\left(
\cos(\hat{\mathbf{z}}_i, \hat{\mathbf{z}}_j) - w_{ij}
\right)^2
\end{equation}
is a similarity loss over connected node pairs that enforces consistency with the realism ontology, and 
\begin{equation}
\mathcal{L}_{\text{rep}} =
\mathbb{E}_{(i,j)\notin E}
\left[
\cos(\mathbf{z}_i, \mathbf{z}_j)^2
\right]
\end{equation}
is a repulsion loss over randomly sampled unconnected pairs that helps avoiding degenerate solutions. 
This ensures that the learned representations not only cluster related concepts together but also separate concepts that are semantically incompatible according to the realism ontology. 

Rather than aggregating embeddings (e.g. pooling), we preserve the full structured representation of size $N \times k$, maintaining trait-specific semantics and relational information encoded by the graph.
Formally, given the node embeddings $\{\mathbf{z}_i\}_{i=1}^N$ produced by the realism GNN,
we define the global realism representation as their ordered collection:
\begin{equation}
    \mathbf{G} = [\mathbf{z}_1, \mathbf{z}_2, \dots, \mathbf{z}_N] \in \mathbb{R}^{N \times k}.
    \label{eq:global-emb}
\end{equation}
This structured representation is used as input to the diffusion model conditioning and the meta-classifier.





\paragraph{Meta-Classifier for Evaluation.}
To quantitatively assess the quality of our realism ontology, the effectiveness of the graph-based embeddings, and the reliability of the VLM-generated trait labels (c.f. Section \ref{sec:experiments}), we further construct a separate meta-classifier
$f_{\text{eval}} : \mathbb{R}^{N \cdot k} \rightarrow \{0,1\}$ to distinguish real from synthetic images based on realism embeddings. Specifically, the meta-classifier operates on embeddings derived from trait probabilities predicted by the supervised MLP heads and refined via graph-based reasoning. Trait probability vectors are propagated through the frozen realism GNN, producing node-level embeddings corresponding to individual realism traits. 
This structured realism embedding is then flattened and input to a shallow MLP meta-classifier trained with a cross-entropy loss to distinguish real from synthetic images. The meta-classifier serves solely as a \textit{diagnostic tool} to quantitatively assess the discriminative power and coherence of the learned realism representation; during evaluation, the trait predictors and realism GNN remain frozen, and only the meta-classifier parameters are updated.


\subsection{Symbolic Planning for Realism-Aware Prompt Generation}
\label{sec:pddl}

To translate realism discrepancies into explicit and controllable editing guidance, we employ symbolic planning using the Planning Domain Definition Language (PDDL). 
We define a PDDL domain in which realism traits are represented as predicates and their causal dependencies are encoded as actions (c.f. Figure~\ref{fig:pipeline}). 
Each action specifies admissible transformations between traits, with preconditions and effects (similar to our ontology-guided knowledge graph) enforcing physical and perceptual consistency—mutually incompatible traits cannot be simultaneously activated, and prerequisite traits must be satisfied before dependent traits can be enabled. 
We use a single, shared PDDL domain across all experiments to encode domain-independent realism constraints, rather than learning symbolic structure from data.

Given a synthetic input image $x_s$ and a target realism specification—either derived from a paired real image $x_t$, an arbitrary reference image, or an explicitly defined realism configuration—we first extract the corresponding trait probability vectors $\mathbf{p}_s = (p_{s,1}, \dots, p_{s,N})$ and $\mathbf{p}_t = (p_{t,1}, \dots, p_{t,N})$ using the supervised trait prediction heads (Section~\ref{sec:traits}). 
These vectors encode the appearance-level realism of the input and desired target states, respectively, and are binarised via a fixed threshold to yield boolean vectors $\mathbf{b}_s, \mathbf{b}_t \in \{0,1\}^N$.

For each image pair, a PDDL problem instance is generated automatically. 
Predicates corresponding to active traits in $\mathbf{b}_s$ define the initial state, while predicates corresponding to $\mathbf{b}_t$ define the goal state. 
To compute the sequence of transformations from the initial to the goal state, we employ the Fast Downward planner \cite{helmert2006fast}.

We adopt a symbolic planner rather than directly flipping boolean traits (a common solution) since realism traits are not independent. 
For example, activating a lighting-related trait may require enabling or disabling dependent shadow or material traits \footnote{If a requested target configuration violates these constraints or represents an infeasible realism state, the planner correctly reports that no valid plan exists.}.

The planner then produces an ordered sequence of symbolic actions that transform the synthetic realism configuration into the target state. 
By integrating these ontology-consistent, causally valid plans, we ensure that subsequent diffusion-based editing respects both perceptual realism and relational dependencies between traits.

\subsection{Ontology-Guided Diffusion Model}
\label{sec:diffusion}

The above symbolic plan is converted into a structured textual prompt by mapping each PDDL action to a concise natural-language instruction (c.f. Figure \ref{fig:pipeline}).
The final prompt preserves action order and semantic intent, yielding interpretable, minimal, and causally grounded editing instructions.
These prompts are embedded using the diffusion model’s text encoder and serve as the linguistic conditioning signal.
Compared to free-form prompts, this representation provides explicit, interpretable, and causally consistent guidance.

We build on an instruction-guided image editing diffusion model (InstructPix2Pix \cite{brooks2023instructpix2pix}) and augment it with realism-aware conditioning derived from symbolic planning and knowledge-graph embeddings.
The diffusion backbone is a conditional UNet \cite{brooks2023instructpix2pix} that predicts noise given a noisy latent, a timestep, and a set of conditioning tokens.


\paragraph{Graph Embedding Conditioning.}
In addition to textual instructions, we condition the diffusion model on the realism structure encoded by our knowledge graph.
Given node-level realism embeddings $\{\mathbf{z}_i\}_{i=1}^N$ produced by the GNN (Section~\ref{sec:gnn}), we treat each node embedding as a separate conditioning token.
Concretely, node embeddings are stacked into a tensor
$\mathbf{Z} \in \mathbb{R}^{B \times N \times d_{\text{KG}}}$,
where $B$ is the batch size and $N$ is the number of realism traits.
These embeddings are then linearly projected to the UNet cross-attention dimension:
$\tilde{\mathbf{Z}} = \mathbf{Z} \mathbf{W}_{\text{KG}}$, where $\mathbf{W}_{\text{KG}} \in \mathbb{R}^{d_{\text{KG}} \times d_{\text{attn}}}.$
To preserve node identity, we add learned positional embeddings $\mathbf{P} \in \mathbb{R}^{1 \times N \times d_{\text{attn}}}$: $\hat{\mathbf{Z}} = \tilde{\mathbf{Z}} + \mathbf{P}$.
The resulting tokens are concatenated with the text embeddings produced by the text encoder, 
$\mathbf{E}_{\text{cond}} = [\mathbf{E}_{\text{text}} ; \hat{\mathbf{Z}}]$,
and passed 
to the cross-attention layers of the UNet.
This allows the model to jointly attend to linguistic instructions and structured realism information at every denoising step.

\paragraph{Diffusion Process with Ontology-Guided Conditioning.}
Let $\mathbf{x}_0$ denote the latent representation of the target image and $\mathbf{x}_t$ a noisy version at timestep $t$.
The diffusion model learns to predict the added noise $\boldsymbol{\epsilon}$ as:
\begin{equation}
\boldsymbol{\epsilon}_\theta =
\epsilon_\theta(\mathbf{x}_t, t, \mathbf{E}_{\text{cond}}),
\end{equation}
where $\mathbf{E}_{\text{cond}}$ includes both symbolic prompt embeddings and realism graph tokens.
Apart from this augmented conditioning, the diffusion process follows the standard denoising diffusion formulation.

\paragraph{Training Objective.}
The diffusion model is trained using the standard denoising diffusion objective. Specifically, given a clean image $\mathbf{x}_0$, a timestep $t$, and Gaussian noise $\boldsymbol{\epsilon} \sim \mathcal{N}(0, I)$, a noisy image $\mathbf{x}_t$ is constructed following the forward diffusion process. The model is trained to predict the added noise:
\begin{equation}
\mathcal{L}_{\text{diff}} =
\mathbb{E}_{\mathbf{x}_0, \boldsymbol{\epsilon}, t}
\left[
\left\|
\boldsymbol{\epsilon} -
\boldsymbol{\epsilon}_\theta(\mathbf{x}_t, t, c)
\right\|_2^2
\right],
\end{equation}
where $c$ denotes the conditioning inputs, including the symbolic prompt and realism graph embeddings.

In addition, we introduce a realism alignment loss to encourage generated images to match the target realism structure. Given a denoised prediction $\hat{\mathbf{x}}_0$, obtained from the model’s noise prediction at timestep $t$, we re-extract its global realism embedding $\mathbf{g}_{\text{gen}}$ using the same frozen trait extractors and realism graph encoder described in Sections~\ref{sec:traits}--\ref{sec:gnn}. This embedding is compared to the target realism embedding $\mathbf{g}_{\text{tgt}}$ via a cosine similarity loss:

\begin{equation}
\mathcal{L}_{\text{KG}}
=
\frac{1}{N}
\sum_{i=1}^{N}
\left(
1 - \cos\!\left(
\mathbf{z}^{(i)}_{\text{gen}},
\mathbf{z}^{(i)}_{\text{tgt}}
\right)
\right),
\end{equation}

Cosine similarity is computed independently between corresponding realism trait embeddings (nodes), with the per-trait losses averaged to form the final loss.
The trait predictors and graph network are kept fixed during diffusion training, so gradients from this auxiliary loss affect only the diffusion model.
The final training objective is:
\begin{equation}
\mathcal{L} =
\mathcal{L}_{\text{diff}} +
\lambda_{\text{KG}} \mathcal{L}_{\text{KG}}.
\end{equation}


\section{Experiments}
\label{sec:experiments}

We evaluate our ontology-guided diffusion framework along two dimensions: (i) whether the GNN on the realism knowledge graph produces meaningful, discriminative embeddings, and (ii) the effectiveness of these embeddings in guiding sim2real image translation. Together, these experiments assess both the quality of the learned realism representations and their utility for image generation.


\subsection{Datasets}
\label{sec:datasets}

We use two complementary categories of data, serving distinct roles in our framework. Both datasets are \emph{only} used for training and are not required at inference time: 
\begin{itemize}[leftmargin = *]
    \item \textbf{Unpaired Synthetic and Real Images (Realism Representation Learning):}
    To train the realism trait classifiers and the realism knowledge graph, we use a collection of \textit{only} $140$ unpaired synthetic and real images drawn from diverse sources.
    Synthetic images are generated using multiple simulation environments, including CARLA ~\cite{dosovitskiy2017carla},
    MuJoCo~\cite{todorov2012mujoco}, Unreal Engine~\cite{unrealengine}, and
    NVIDIA Isaac Sim~\cite{makoviychuk2021isaac}, covering a wide range of object categories and scene layouts.
    Real images are collected from large-scale real-world datasets and internet imagery (e.g., COCO ~\cite{lin2014microsoft}), without enforcing image-level correspondence to synthetic samples.
    This unpaired setting exposes the model to systematic appearance discrepancies between simulation and reality, such as differences in illumination, material reflectance, sensor noise, and optical artefacts, while remaining agnostic to scene-specific alignment.


    \item \textbf{Paired Sim2Real Images (Diffusion Model Training):}
    For training the diffusion-based image editing model, we use paired synthetic–real image datasets where each synthetic image has a corresponding real counterpart with matched scene geometry and semantics.
    In particular, we use the Virtual KITTI 2 dataset \cite{cabon2020vkitti2,gaidon2016virtual} paired with the real KITTI dataset \cite{Geiger2012CVPR}, which provides high-quality synthetic renderings aligned with real-world driving scenes.
    In addition, we include a proprietary lego dataset (Figure \ref{fig:qualitative}) consisting of 113 curated synthetic images and their corresponding real images. 
\end{itemize}

\subsection{Evaluation of Realism Representation}
\label{sec:exp-representation}

We first evaluate whether the proposed realism graph embeddings capture meaningful structure that distinguishes real from synthetic imagery.
Using the trait-based realism representation defined in Section~\ref{sec:traits}, we obtain a realism embedding for each image by propagating predicted trait probabilities through the frozen realism GNN.
Rather than collapsing node embeddings via mean pooling, we retain the full set of node-level embeddings, resulting in a structured representation of size $N \times d$ (with $N$ traits and embedding dimension $d$).
This preserves trait-specific semantics and avoids destroying relational information encoded by the graph.
Our lightweight meta-classifier is then trained to predict whether an image originates from a real or synthetic domain based on this structured realism embedding.
During this evaluation, the realism GNN and trait predictors are kept fixed, and only the meta-classifier is trained.
This protocol ensures that performance reflects the quality of the learned realism representation rather than classifier capacity.

We compare against: (i) a linear classifier trained on frozen CLIP embeddings (\textit{CLIP features}), (ii) a classifier trained on raw trait probabilities (\textit{Trait meta-classifier}), and (iii) a CNN classifier trained from scratch (\textit{End-to-end CNN}).

We report classification accuracy and the area under the receiver operating characteristic curve (ROC–AUC) on a held-out test set.

\subsection{Evaluation of Sim2Real Image Translation}
\label{sec:exp-sim2real}

Next, we evaluate our ontology-guided diffusion for sim2real image translation.
Given a synthetic input image, we generate an edited image using an InstructPix2Pix diffusion model conditioned on:
(i) a structured instruction prompt derived from symbolic planning, and
(ii) node-level realism embeddings injected via cross-attention.

We compare against the state-of-the-art: (i) \textit{InstructPix2Pix} \cite{brooks2023instructpix2pix} with static hand-designed realism prompt, and (ii) \textit{ControlNet} \cite{zhang2023controlnet} the diffusion-based sim2real baseline with depth maps as structural conditioning.

We evaluate sim-to-real translation using complementary semantic and perceptual measures: (i) \textit{Trait distance} (\textit{TraitDist}): Euclidean distance between predicted continuous trait vectors of the generated image and the real reference. This directly measures semantic alignment with real-world attributes, (ii) \textit{Perceptual similarity} (\textit{LPIPS}): Learned perceptual distance between the generated and real images, capturing human-aligned visual similarity beyond pixel-wise differences, and (iii) \textit{Structural similarity} (\textit{SSIM}): Structural consistency between generated and real images, indicating preservation of scene layout and object geometry (c.f. Appendix~\ref{app:metrics} for formal definitions).


\section{Results}
\label{sec:results}

\begin{figure*}[t]
    \centering
    \setlength{\tabcolsep}{2pt}

    \begin{tabular}{cccccc}
     &
        \textbf{Synthetic} &
        \textbf{InstructPix2Pix} &
        \textbf{ControlNet} &
        \textbf{Ours} &
        \textbf{Real Reference} \\

        \rotatebox{90}{\textbf{KITTI}} & \includegraphics[width=0.15\linewidth]{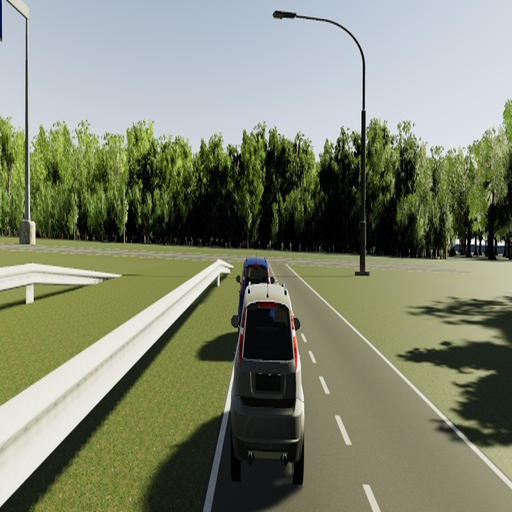} &
        \includegraphics[width=0.15\linewidth]{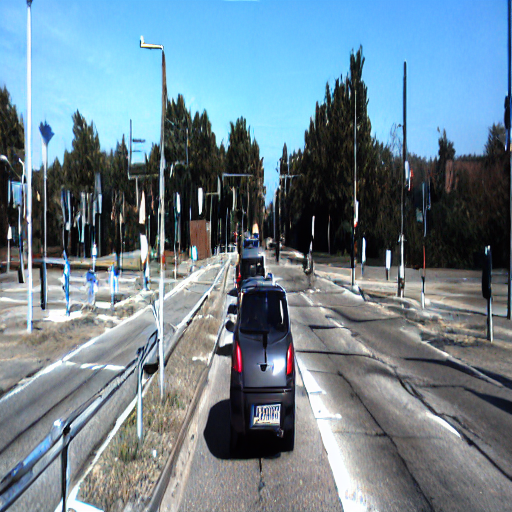} &
        \includegraphics[width=0.15\linewidth]{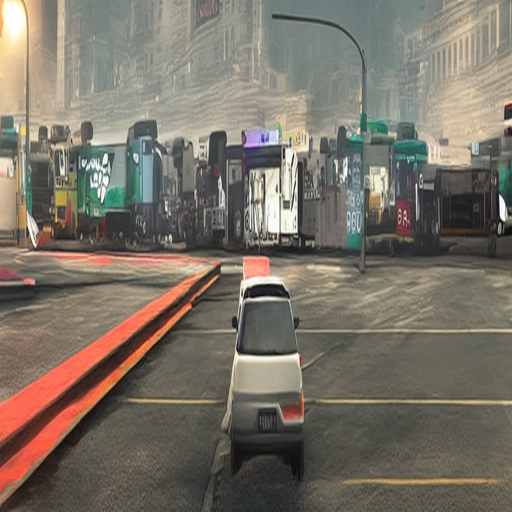} &
        \includegraphics[width=0.15\linewidth]{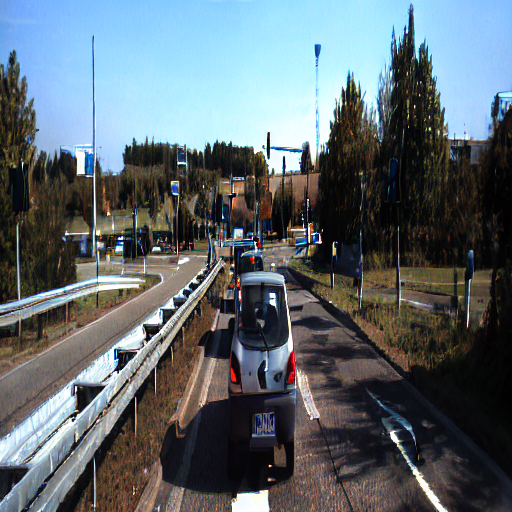} &
        \includegraphics[width=0.15\linewidth]{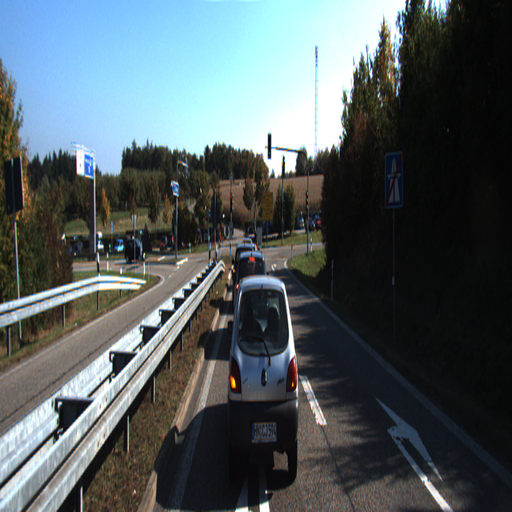} \\

        \rotatebox{90}{\textbf{LEGO}} & \includegraphics[width=0.15\linewidth]{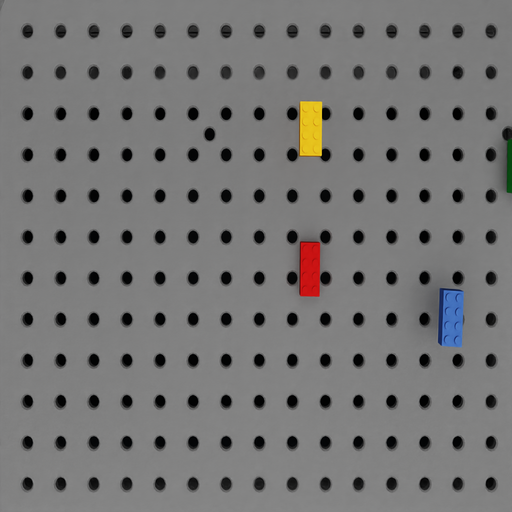} &
        \includegraphics[width=0.15\linewidth]{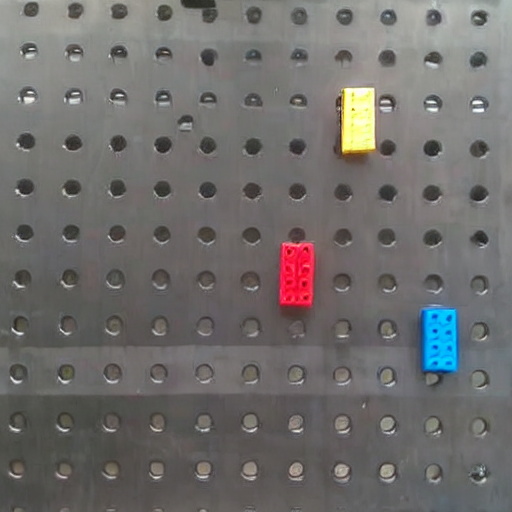} &
       
        \includegraphics[width=0.15\linewidth]{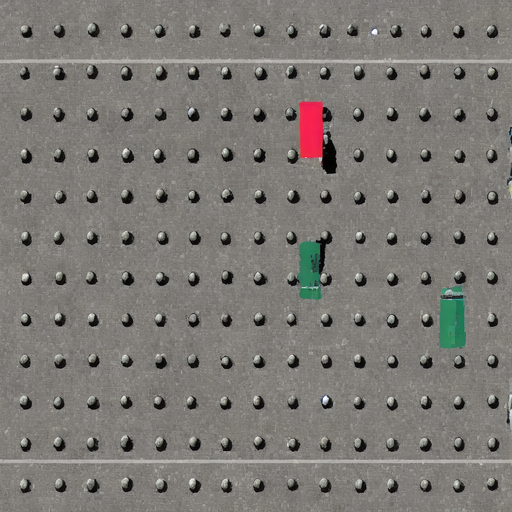} &
        \includegraphics[width=0.15\linewidth]{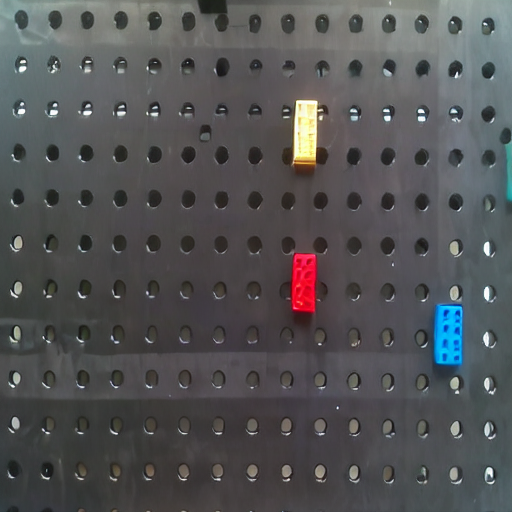} &
        \includegraphics[width=0.15\linewidth]{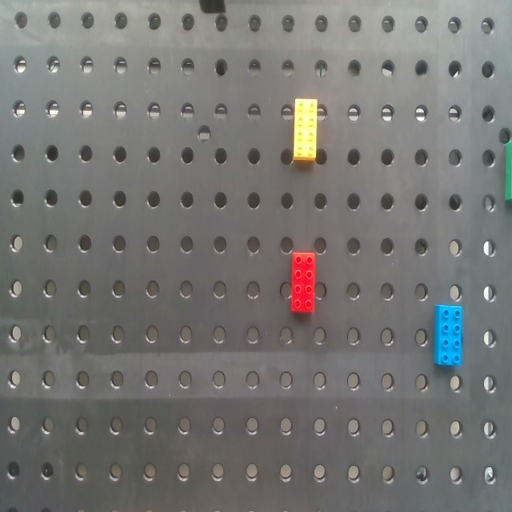}
    \end{tabular}

    \caption{
    Qualitative sim-to-real results. 
    }
    \label{fig:qualitative}
\end{figure*}

We first evaluate the quality of the learned realism representations, followed by results on sim2real image translation.

\subsection{Realism Representation Performance}
\label{sec:results-representation}


Table~\ref{tab:meta_classifier} reports meta-classifier performance across different input representations. All results are reported on a held-out test set with a fixed decision threshold of 0.9 for
meta-classifiers.
Across all metrics, classifiers trained on GNN-based realism embeddings significantly outperform baselines that rely on unstructured visual features or independent trait predictions. 
This demonstrates that explicitly modeling interactions between realism traits—such as supportive and opposing relationships—yields representations that better capture global realism coherence.
Table \ref{tab:meta_classifier} further demonstrates that incorporating the realism ontology via graph propagation yields substantially more discriminative embeddings than unstructured trait vectors or raw visual features, highlighting the importance of modeling realism coherence explicitly.
Notably, using trait probabilities without graph propagation results in substantially lower performance, despite identical supervision. 
This confirms that realism cannot be reliably inferred by treating appearance cues independently. 
By contrast, the realism ontology allows the model to resolve conflicting visual evidence.

\begin{table}[h]
  \centering
  \caption{Real vs. synthetic classification performance.}
  \label{tab:meta_classifier}
  \resizebox{\linewidth}{!}{%
  \begin{tabular}{lcc}
    \toprule
    Method & Accuracy (\%) & ROC--AUC \\
    \midrule
    CLIP features & 59.6 & 0.88 \\
    End-to-end CNN & 76.5 & 0.68 \\
    Traits meta-classifier & 97.0 & 0.90 \\
    \textbf{Ours (Traits + GNN meta-classifier)} & \textbf{98.4} & \textbf{0.99} \\
    \bottomrule
  \end{tabular}
  }
\end{table}

\subsection{Sim2Real Image Translation Performance}
\label{sec:results-sim2real}

Quantitatively, as reported in Table~\ref{tab:sim2real_quant}, \methodName{} achieves the highest realism classification accuracy and the lowest distributional distance to real images. Conditioning the diffusion model on both symbolic transformation plans and realism graph embeddings consistently outperforms baseline approaches, including InstructPix2Pix (static prompts) and ControlNet (static prompts augmented with depth images).

\begin{table}[h]
\caption{Quantitative sim-to-real image translation performance.
}
\label{tab:sim2real_quant}
\centering
\begin{small}
\begin{tabular}{lccc}
\toprule
\textbf{Method} &
\textbf{TraitDist} $\downarrow$ &
\textbf{LPIPS} $\downarrow$ &
\textbf{SSIM} $\uparrow$ \\
\midrule
I-Pix2Pix &
0.702 $\pm$ 0.413 &
0.315 $\pm$ 0.247 &
0.534 $\pm$ 0.332 \\
ControlNet &
3.533 $\pm$ 0.717 &
0.530 $\pm$ 0.116 &
0.286 $\pm$ 0.150 \\

\textbf{Ours} &
\textbf{0.593 $\pm$ 0.509} &
\textbf{0.275 $\pm$ 0.220} &
\textbf{0.562 $\pm$ 0.307} \\
\bottomrule
\end{tabular}
\end{small}
\end{table}

Qualitatively, figure~\ref{fig:qualitative} shows representative examples of sim2real translation. 
Overall, \methodName{} produces images with more coherent lighting gradients, realistic material imperfections, and improved scene-level consistency, while preserving semantic structure and content (c.f. Appendix \ref{apx:qualitative} for more qualitative examples). 
In contrast, baseline methods frequently introduce inconsistent shadows, overly smooth textures, or fail to correct key realism deficiencies.

\subsection{Ablation Analysis}
\label{sec:results-ablation}

We perform ablation studies to isolate the contribution of each component of the proposed framework. As shown in Table~\ref{tab:ablation}, removing the realism graph or replacing symbolic planning with free-form prompts leads to consistent performance degradation, particularly in scenarios requiring coordinated changes across multiple appearance traits. Omitting the realism alignment loss results in less stable realism transfer, highlighting the importance of enforcing structural consistency during diffusion training. 
qualitative results are provided in Appendix~\ref{apx:ablation}. 

\begin{table}[h]
\caption{Ablation study on sim-to-real image translation.}
\label{tab:ablation}
\centering
\resizebox{\linewidth}{!}{%
\begin{tabular}{lccc}
\toprule
\textbf{Configuration} &
\textbf{TraitDist} $\downarrow$ &
\textbf{LPIPS} $\downarrow$ &
\textbf{SSIM} $\uparrow$ \\
\midrule
\textbf{Full model (ours)} &
\textbf{0.693} &
\textbf{0.275} &
\textbf{0.562} \\
\midrule
-- w/o realism graph &
0.842 &
0.312 &
0.524 \\
-- w/o symbolic planning &
0.781 &
0.298 &
0.538 \\
-- w/o KG alignment loss &
0.756 &
0.287 &
0.546 \\
\bottomrule
\end{tabular}

}
\end{table}





\section{Conclusion}

We introduced \methodName{}, a neuro-symbolic framework for zero-shot sim2real image translation that explicitly models visual realism as structured knowledge rather than an implicit latent. By decomposing realism into interpretable appearance traits, encoding their interactions in a knowledge graph, and using symbolic planning to guide diffusion-based image editing, our approach enables causally consistent and interpretable sim2real transformations. Empirically, realism graph embeddings better distinguish real from synthetic imagery, and ontology-guided diffusion produces more coherent results than the SOTA baselines. Crucially, training requires only a small dataset to learn realism traits, and inference is fully zero-shot, making the approach highly data-efficient and well suited to robotics and simulation-heavy domains with scarce paired data.

\section*{Impact Statement}

This paper presents research aimed at advancing the field of visual learning systems by addressing the sim2real gap, a challenge with direct implications for robotics, autonomous systems, and embodied AI. 
While such technologies may have broad societal consequences, the proposed approach is not associated with negative societal impact; on the contrary, it is expected to provide benefits in terms of privacy, ethics, and sustainability. 
By enabling more reliable transfer from synthetic data to real-world environments, the method has the potential to reduce the cost, environmental footprint, and privacy risks associated with large-scale real-world data collection and manual annotation. 


\bibliography{references}
\bibliographystyle{icml2026}


\newpage
\appendix
\onecolumn

\section{Appendix}

\subsection{Realism Trait Ontology}
\label{app:ontology}

The realism ontology consists of a fixed set of appearance-level traits covering lighting, shadows, material properties, geometric cues, optical and sensor artifacts, color behavior, and scene consistency. Each trait is represented as a binary or probabilistic variable and corresponds to a node in the realism knowledge graph.

The ontology and its relationships are manually specified based on prior work in computer graphics, visual perception, and image forensics. Importantly, the ontology is static and shared across all datasets and experiments; no ontology structure or relationships are learned from data. 

\begin{table*}[htbp]
\centering
\small
\setlength{\tabcolsep}{6pt}
\renewcommand{\arraystretch}{1.2}
\begin{tabular}{p{4.5cm} c p{4.5cm} c p{6.2cm}}
\hline
\textbf{Source Trait (Observable / Inferable)} & & \textbf{Target Trait} & \textbf{Rel.} & \textbf{Supporting Evidence} \\
\hline
\multicolumn{5}{l}{\textbf{Visually Observable and Scene-Inferable Trait Relationships}} \\[4pt]

Lighting: Uniform
& $\Rightarrow$ 
& Shadows: Present
& opposes 
& Diffuse illumination eliminates cast shadows as spatial cues \cite{ferwerda2003three}. \\

Shadows: Present
& $\Rightarrow$ 
& Scene Consistency: Object Interaction
& supports 
& Cast shadows convey spatial relationships between objects \cite{ferwerda2003three}. \\

Optical \& Sensor Artifacts: Chromatic Aberration
& $\Rightarrow$ 
& Edge \& Geometry: Perfect Geometry
& opposes 
& Color fringing along high-contrast edges disrupts geometric precision \cite{carlson2018modeling}. \\

Optical \& Sensor Artifacts: Blur or Depth of Field
& $\Rightarrow$ 
& Optical \& Sensor Artifacts: Noise Present
& supports 
& Blur and noise commonly co-occur as visible image degradations \cite{carlson2018modeling}. \\

Optical \& Sensor Artifacts: Noise Present
& $\Rightarrow$ 
& Optical \& Sensor Artifacts: Compression Artifacts
& supports 
& Noise interacts with compression, producing visible artifacts \cite{elmquist2021modeling}. \\

Edge \& Geometry: Lens Distortion Present
& $\Rightarrow$ 
& Optical \& Sensor Artifacts: Chromatic Aberration
& supports 
& Lens-induced artifacts such as distortion and chromatic aberration frequently co-occur \cite{elmquist2021modeling}. \\

Optical \& Sensor Artifacts: Vignetting
& $\Rightarrow$ 
& Optical \& Sensor Artifacts: Noise Present
& supports 
& Vignetting is often accompanied by increased visible noise in darker regions \cite{elmquist2021modeling}. \\

Optical \& Sensor Artifacts: Lens Flare
& $\Rightarrow$ 
& Scene Consistency: Environmental Integration
& opposes 
& Ghosting artifacts disrupt physical consistency between objects and environment \cite{elmquist2021modeling}. \\

Color \& Reflectivity: Oversaturation
& $\Rightarrow$ 
& Scene Consistency: Object Interaction
& opposes 
& Over-saturation degrades object appearance and perceptual consistency \cite{carlson2018modeling}. \\

Scene Consistency: Object Interaction
& $\Rightarrow$ 
& Scene Consistency: Realistic Scatter
& supports 
& Shadows and inter-reflections visually connect objects within a scene \cite{ferwerda2003three}. \\

Scene Consistency: Realistic Scatter
& $\Rightarrow$ 
& Scene Consistency: Environmental Integration
& supports 
& Global light scattering unifies object and environment appearance \cite{elmquist2021modeling}. \\

\hline
\end{tabular}
\caption{A part of the paper-supported trait--trait relationships restricted to properties that are visually observable or inferable from a single image by a strong vision--language model. Trait names correspond directly to canonical knowledge graph identifiers.}
\label{tab:trait_relations_vlm}
\end{table*}




\subsection{Knowledge Graph Construction}
\label{app:kg}

The realism knowledge graph encodes signed relationships between traits. Supportive relationships are assigned positive edge weights, while opposing relationships are assigned negative weights. Edge weights are normalized to a fixed range for numerical stability, and bidirectional edges are added to enable message passing.

Self-loops are included for all nodes to preserve isolated traits and ensure stable graph propagation.

\begin{figure}[h]
\centering
\includegraphics[width=0.9\linewidth]{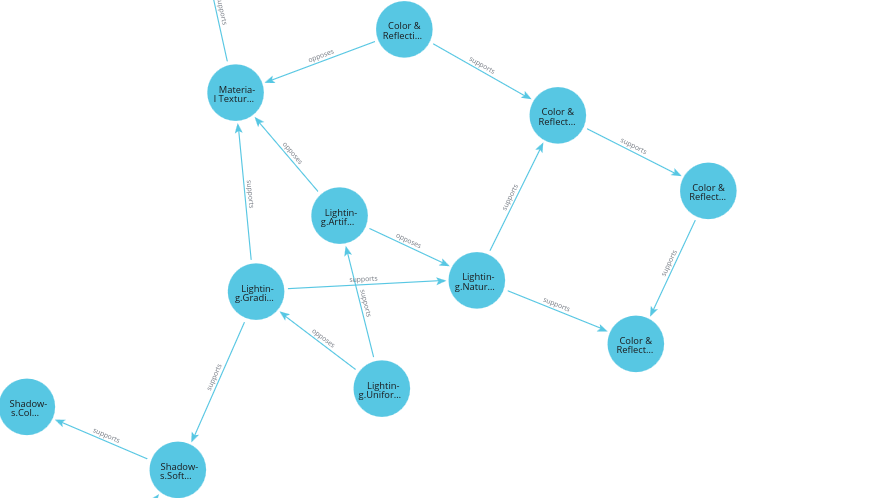}
\caption{Visualization of part of the realism knowledge graph generated using Neo4j Browser. Nodes correspond to realism traits, while signed edges encode supportive (positive) and opposing (negative) relationships derived from graphics and perception literature.}
\label{fig:kg}
\end{figure}

\newpage
\subsection{Graph Neural Network Architecture and Training}
\label{app:gnn}

The GNN uses a two-layer GraphSAGE architecture. Node features are initialized from trait probabilities predicted by the supervised MLP heads. The GNN is trained using a signed similarity loss that aligns embeddings of supportive traits and separates embeddings of opposing traits, combined with a repulsion regularizer applied to randomly sampled unconnected node pairs.

Once trained, the GNN is frozen and used to generate realism embeddings for all downstream tasks, including evaluation and diffusion conditioning.

\subsection{Meta-Classifier for Realism Evaluation}
\label{app:meta}

To quantitatively evaluate the quality of the realism embeddings, we train a lightweight meta-classifier on GNN embeddings to distinguish real from synthetic images. The GNN is frozen during this stage, ensuring that classification performance reflects the quality of the learned realism representation rather than classifier capacity.

This meta-classifier is used solely for benchmarking and is not involved in diffusion training or inference.

\subsection{Diffusion Model Modifications}
\label{app:diffusion}

The diffusion model is based on an InstructPix2Pix architecture. To enable realism-aware conditioning, node-level realism embeddings are projected to the cross-attention dimension and concatenated with text encoder hidden states. Positional embeddings are added to preserve node identity.

During training, a realism alignment loss encourages generated images to match target realism embeddings extracted using frozen trait and GNN encoders.

\subsection{Implementation Details}
All models are trained using Adam optimizers with fixed learning rates.
The CLIP encoder is frozen throughout all experiments.
Graph structure and symbolic planning rules are fixed across datasets.
Additional implementation details are provided in the Appendix. 

\subsection{Evaluation Metrics}
\label{app:metrics}
\paragraph{Trait Distance (TraitDist).}
Let $\mathbf{p}_{\text{gen}} \in [0,1]^N$ and $\mathbf{p}_{\text{real}} \in [0,1]^N$ denote the continuous realism trait probability vectors
predicted by the trained trait classifiers for the generated image and the real reference image, respectively.
Trait distance is defined as the Euclidean distance between these vectors:
\begin{equation}
\mathrm{TraitDist}(\mathbf{p}_{\text{gen}}, \mathbf{p}_{\text{real}}) =
\left\| \mathbf{p}_{\text{gen}} - \mathbf{p}_{\text{real}} \right\|_2.
\end{equation}
Lower values indicate closer alignment in predicted realism attributes.
\paragraph{Perceptual Similarity (LPIPS).}
Perceptual similarity is measured using the Learned Perceptual Image Patch Similarity (LPIPS) metric.
Given a generated image $x_{\text{gen}}$ and a real reference image $x_{\text{real}}$, LPIPS computes the
distance between deep feature activations extracted from a pretrained convolutional network:
\begin{equation}
\mathrm{LPIPS}(x_{\text{gen}}, x_{\text{real}}) =
\sum_{l} \left\| \phi_l(x_{\text{gen}}) - \phi_l(x_{\text{real}}) \right\|_2^2,
\end{equation}
where $\phi_l(\cdot)$ denotes the feature activations at layer $l$.
Lower values correspond to higher perceptual similarity.
\paragraph{Structural Similarity (SSIM).}
Structural similarity is measured using the SSIM index, which compares local luminance, contrast,
and structure between two images.
Given image patches $x$ and $y$, SSIM is defined as:
\begin{equation}
\mathrm{SSIM}(x,y) =
\frac{(2\mu_x\mu_y + C_1)(2\sigma_{xy} + C_2)}
{(\mu_x^2 + \mu_y^2 + C_1)(\sigma_x^2 + \sigma_y^2 + C_2)},
\end{equation}
where $\mu_x, \mu_y$ are mean intensities, $\sigma_x^2, \sigma_y^2$ are variances,
$\sigma_{xy}$ is covariance, and $C_1, C_2$ are stabilizing constants.
Higher values indicate stronger structural similarity.




\subsection{Further Qualitative Results}\label{apx:qualitative}

\begin{figure*}[t]
    \centering
    \setlength{\tabcolsep}{2pt}

    \begin{tabular}{cccccc}
     &
        \textbf{Synthetic} &
        \textbf{InstructPix2Pix} &
        \textbf{ControlNet} &
        \textbf{Ours} &
        \textbf{Real Reference} \\

        \rotatebox{90}{\textbf{KITTI}} & \includegraphics[width=0.15\linewidth]{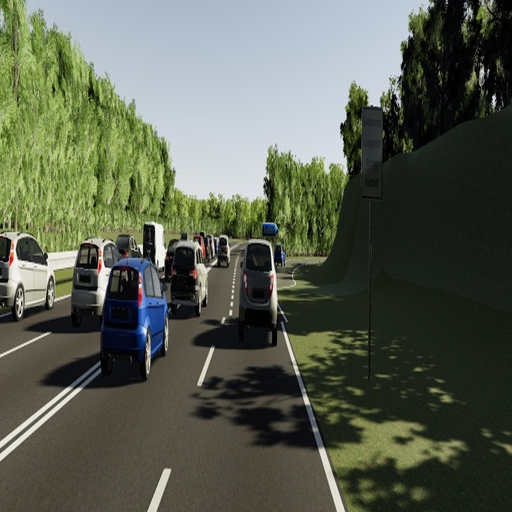} &
        \includegraphics[width=0.15\linewidth]{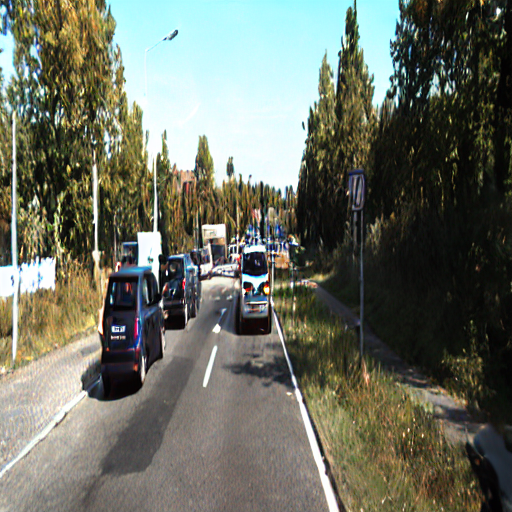} &
        \includegraphics[width=0.15\linewidth]{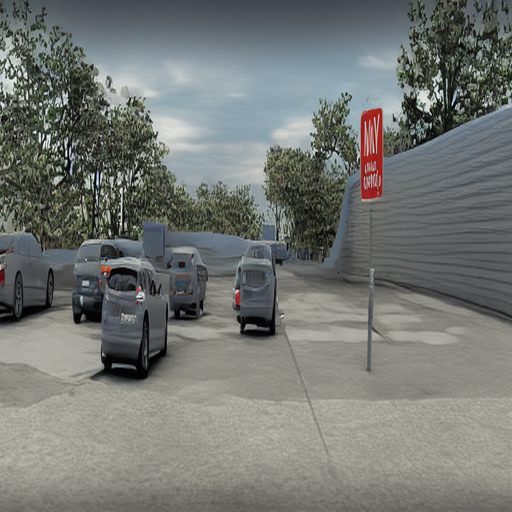} &
        \includegraphics[width=0.15\linewidth]{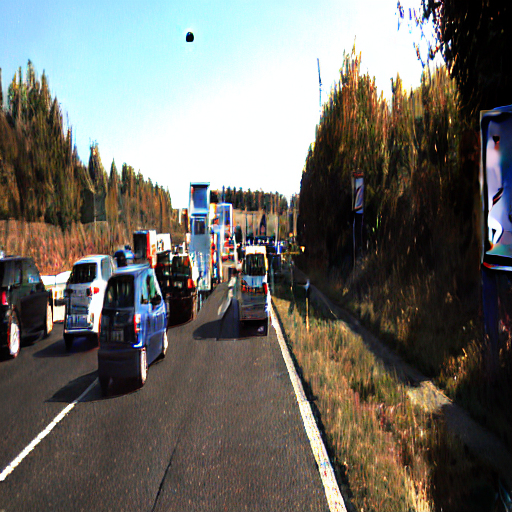} &
        \includegraphics[width=0.15\linewidth]{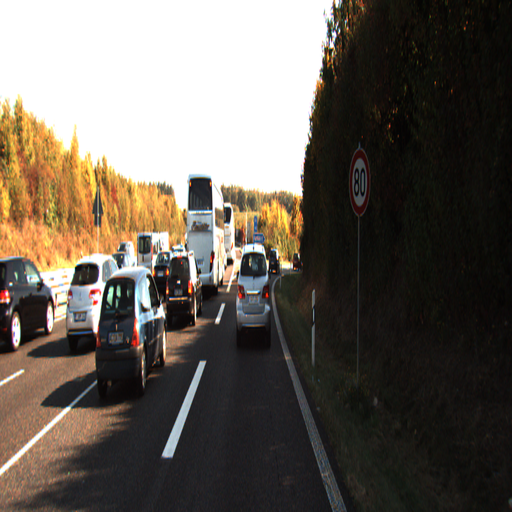} \\

        \rotatebox{90}{\textbf{LEGO}} & \includegraphics[width=0.15\linewidth]{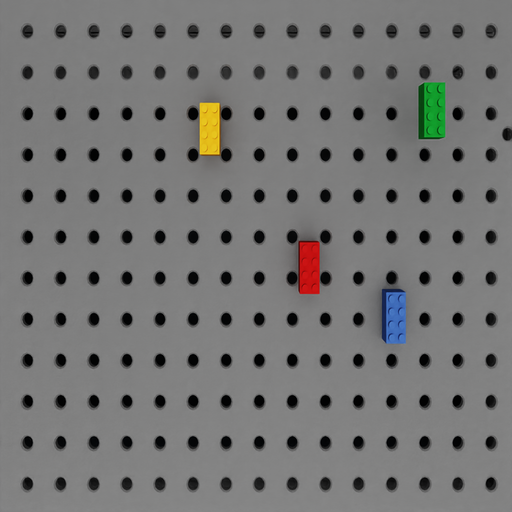} &
        \includegraphics[width=0.15\linewidth]{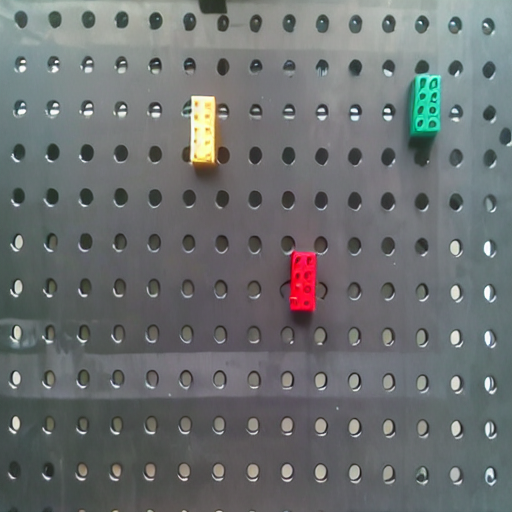} &
       
        \includegraphics[width=0.15\linewidth]{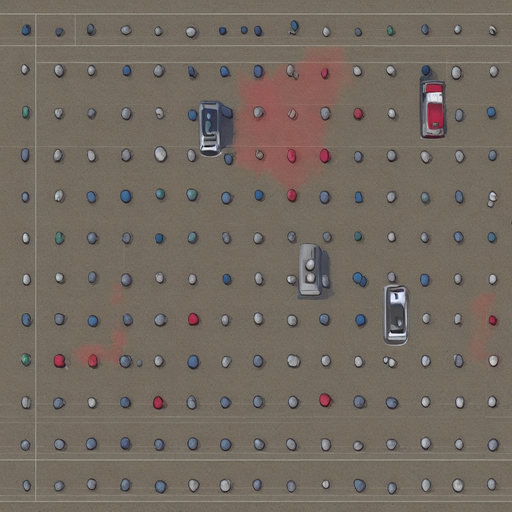} &
        \includegraphics[width=0.15\linewidth]{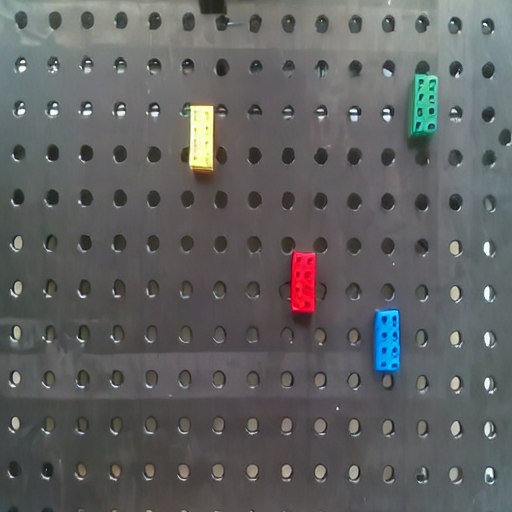} &
        \includegraphics[width=0.15\linewidth]{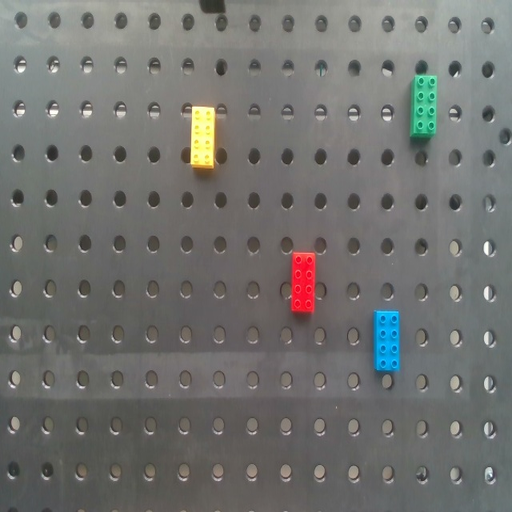}
    \end{tabular}

    \caption{
    Qualitative sim-to-real results. 
    }
    \label{fig:apx-qualitative}
\end{figure*}

As shown in Figure \ref{fig:apx-qualitative}, \methodName{} generates images exhibiting more coherent and physically plausible lighting gradients, realistic material imperfections such as surface roughness and specular variations, and improved scene-level consistency, including coherent interactions between shadows, reflections, and object surfaces. Importantly, these visual enhancements are achieved while preserving the underlying semantic structure and content of the scene, ensuring that object identities, spatial relationships, and compositional layout remain faithful to the input. This combination of perceptual realism and semantic fidelity distinguishes OGD from baseline diffusion approaches that may improve appearance at the cost of structural consistency.

\newpage
\subsection{Ablation Qualitative Results}\label{apx:ablation}
\begin{figure*}[t]
    \centering
    \setlength{\tabcolsep}{2pt}

    \begin{tabular}{cccccc}
        \textbf{Synthetic} &
        \textbf{PDDL w/o KG} &
        \textbf{KG w/o PDDL} &
        \textbf{KG w/o Loss} &
        \textbf{ours KG and PDDL} &
        \textbf{Real Reference} \\

        \includegraphics[width=0.15\linewidth]{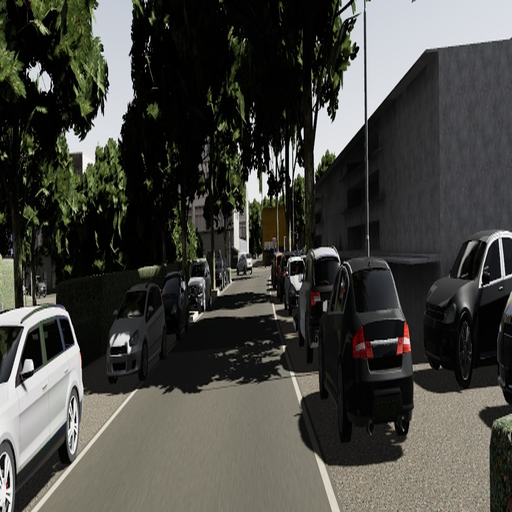} &
        \includegraphics[width=0.15\linewidth]{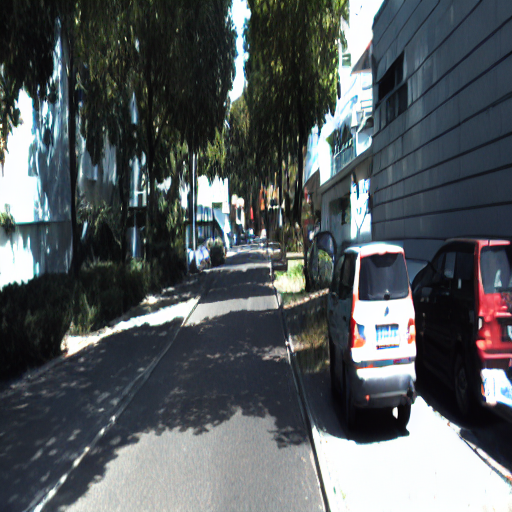} &
        \includegraphics[width=0.15\linewidth]{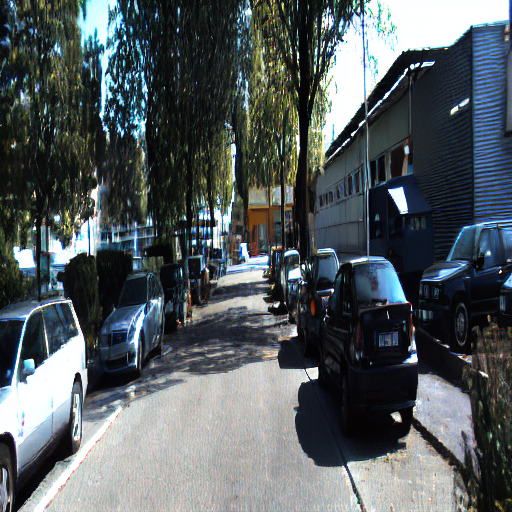} &
        \includegraphics[width=0.15\linewidth]{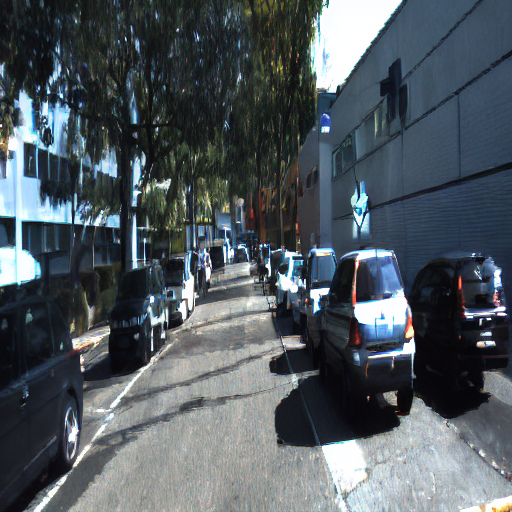} &
        \includegraphics[width=0.15\linewidth]{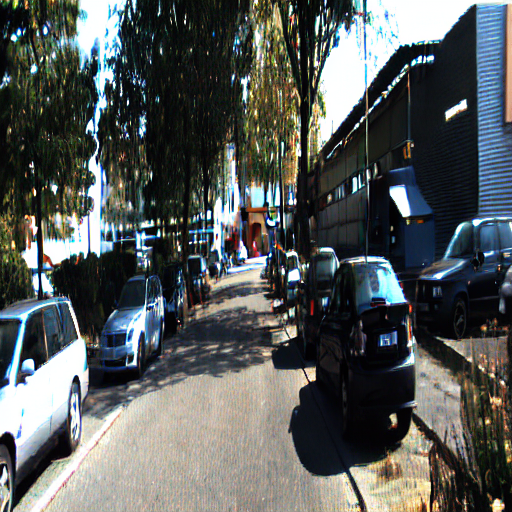} &
        \includegraphics[width=0.15\linewidth]{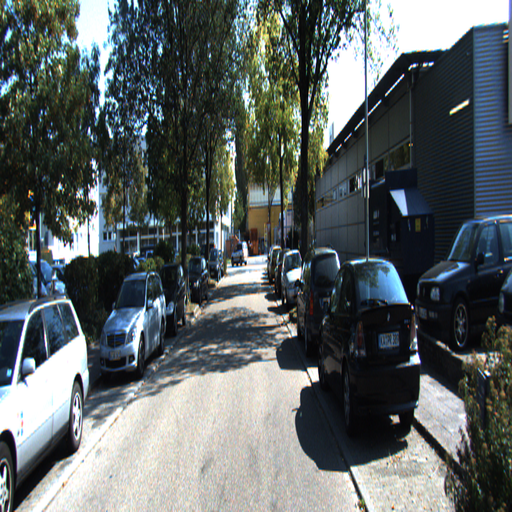} \\
    \end{tabular}

    \caption{Qualitative sim-to-real results on ablation study}
    \label{fig:ablation-qualitative}
\end{figure*}

Qualitative Sim2real results for the ablation study are shown in Figure~\ref{fig:ablation-qualitative}. “PDDL w/o KG” uses symbolic planning–derived text prompts without graph embedding conditioning; “KG w/o PDDL” conditions on realism graph embeddings without symbolic planning, and use A static prompt E.g "Make synthetic image more realistic"; “KG w/o Loss” injects graph embeddings but omits the realism alignment loss with pddl prompt. The full model combines both KG conditioning and symbolic planning.

\end{document}